\documentclass[conference]{IEEEtran}
\IEEEoverridecommandlockouts
\usepackage{cite}
\usepackage{amsmath,amssymb,amsfonts}
\usepackage{algorithmic}
\usepackage{graphicx}
\usepackage{textcomp}
\usepackage{xcolor}
\def\BibTeX{{\rm B\kern-.05em{\sc i\kern-.025em b}\kern-.08em
    T\kern-.1667em\lower.7ex\hbox{E}\kern-.125emX}}
\begin{document}

\title{Detecting Abnormal User Feedback Patterns through Temporal Sentiment Aggregation\\
}

\author{\IEEEauthorblockN{Yalun Qi\textsuperscript{*}}
\IEEEauthorblockA{\textit{Khoury College of Computer Science} \\
\textit{Northeastern University}\\
Boston, United States \\
qi.yal@northeastern.edu}

\and
\IEEEauthorblockN{Sichen Zhao}
\IEEEauthorblockA{\textit{College of Engineering} \\
\textit{Northeastern University}\\
Boston, United States \\
zhao.siche@northeastern.edu}
\and
\IEEEauthorblockN{Zhiming Xue}
\IEEEauthorblockA{\textit{College of Engineering} \\
\textit{Northeastern University}\\
Boston, United States \\
xue.zh@northeastern.edu}
\and
\IEEEauthorblockN{Xianling Zeng}
\IEEEauthorblockA{\textit{College of Engineering} \\
\textit{Northeastern University}\\
Boston, United States \\
zeng.xian@northeastern.edu}
\and
\IEEEauthorblockN{Zihan Yu}
\IEEEauthorblockA{\textit{College of Professional Studies} \\
\textit{Northeastern University}\\
Boston, United States \\
yu.zihan1@northeastern.edu}
}

\maketitle

\begin{abstract}
In many real-world applications, such as customer feedback monitoring, brand reputation management, and product health tracking, understanding the temporal dynamics of user sentiment is crucial for early detection of anomalous events such as malicious review campaigns or sudden declines in user satisfaction. Traditional sentiment analysis methods focus on individual text classification, which is insufficient to capture collective behavioral shifts over time due to inherent noise and class imbalance in short user comments. In this work, we propose a temporal sentiment aggregation framework that leverages pretrained transformer-based language models to extract per-comment sentiment signals and aggregates them into time-window-level scores. Significant downward shifts in these aggregated scores are interpreted as potential anomalies in user feedback patterns. We adopt RoBERTa as our core semantic feature extractor and demonstrate, through empirical evaluation on real social media data, that the aggregated sentiment scores reveal meaningful trends and support effective anomaly detection. Experiments on real-world social media data demonstrate that our method successfully identifies statistically significant sentiment drops that correspond to coherent complaint patterns, providing an effective and interpretable solution for feedback anomaly monitoring.

\end{abstract}


\section{Introduction}

User-generated content on social media platforms, review systems, and online forums has become a critical source of insight for understanding public opinion and collective behavior. Sentiment analysis, also known as opinion mining, aims to identify and categorize expressed emotions or attitudes within textual data \cite{bordoloi2023}. Over the past decade, advances in natural language processing and deep learning have significantly improved per-instance sentiment classification performance.

However, real-world monitoring scenarios extend beyond individual comment classification. In domains such as brand reputation management, service quality tracking, and crisis detection, stakeholders are primarily concerned with detecting abnormal shifts in collective sentiment over time. Wang et al. \cite{wang2014} demonstrate that abnormal events in social media streams often manifest as sudden sentiment changes embedded within large volumes of text. This highlights the importance of integrating sentiment analysis with anomaly detection mechanisms.

Time-series anomaly detection (TSAD) has received substantial attention in recent years, particularly with the rise of deep learning models capable of modeling complex temporal dependencies \cite{darban2023, huang2025}. These approaches typically operate on structured numerical signals. In contrast, sentiment-derived signals originate from unstructured text and inherit uncertainty from language model predictions. Directly applying temporal anomaly detection techniques to raw sentiment classifications may therefore amplify noise rather than reveal meaningful patterns.

Recent research across multiple domains suggests a broader methodological principle: downstream effectiveness is often governed more by objective alignment than architectural complexity. In fraud detection under extreme imbalance, Sun et al. \cite{sun2025} show that models with similar AUC values can exhibit drastically different minority-class performance depending on the optimization objective. Similarly, in reinforcement-learning-based option hedging, Chen et al. \cite{chen2026} demonstrate that optimizing static fitting accuracy does not necessarily yield superior dynamic risk performance. These findings emphasize the distinction between static per-instance metrics and task-aligned operational objectives.

Motivated by these insights, we argue that maximizing individual sentiment classification accuracy is not equivalent to effective anomaly detection in temporal feedback streams. Instead of increasing architectural complexity to jointly model text and time, we propose a modular framework that emphasizes signal stabilization and objective alignment. Specifically, we employ a pretrained transformer-based model to generate per-comment sentiment scores, then aggregate these scores into window-level signals. Significant changes in aggregated sentiment are interpreted as potential anomalies.

Our approach is guided by three principles. First, aggregation mitigates prediction noise without modifying the base language model. Second, change-based detection prioritizes relative shifts over absolute polarity, improving interpretability. Third, the modular design decouples semantic representation from anomaly detection, enabling flexible system deployment.

The main contributions of this paper are as follows:

\begin{itemize}
\item We propose a temporal sentiment aggregation framework that transforms noisy per-comment predictions into stable time-series signals for anomaly monitoring.
\item We introduce a change-based anomaly detection mechanism aligned with operational objectives rather than static classification metrics.
\item We empirically demonstrate that detected sentiment drops correspond to coherent complaint patterns, validating the effectiveness of the proposed approach.
\end{itemize}

\section{Related Work}

\subsection{Sentiment Analysis Frameworks}

Sentiment analysis has evolved from rule-based systems to machine-learning and transformer-based approaches capable of contextual semantic modeling \cite{bordoloi2023}. Modern sentiment classifiers typically frame the task as a polarity classification problem, assigning positive, neutral, or negative labels to textual instances. While high classification accuracy has been achieved in benchmark datasets, most studies evaluate performance under static, instance-level metrics.

\subsection{Sentiment-Based Anomaly Detection}

The integration of sentiment analysis with anomaly detection has been explored in social media contexts. Wang et al. \cite{wang2014} propose detecting abnormal events through enhanced sentiment analysis on Twitter data, demonstrating that sudden shifts in sentiment distribution can correspond to meaningful real-world events. However, such approaches often rely directly on per-instance sentiment outputs without explicitly addressing the instability introduced by short-text noise and class imbalance.

\subsection{Deep Learning for Time-Series Anomaly Detection}

Time-series anomaly detection has become increasingly important across domains including finance, healthcare, and IoT systems. Recent surveys categorize deep TSAD methods into forecasting-based, reconstruction-based, representation-based, and hybrid approaches \cite{darban2023}. Deep neural networks have demonstrated strong capability in modeling high-dimensional temporal dependencies and nonlinear dynamics \cite{huang2025}. 

Nevertheless, most TSAD frameworks assume structured numerical signals. When the underlying signal originates from semantic predictions derived from unstructured text, uncertainty propagation becomes a critical challenge. Stabilizing such signals before applying temporal anomaly detection is therefore essential.

\subsection{Objective Alignment and Representation Robustness}

Beyond anomaly detection, recent work emphasizes the importance of aligning optimization objectives with downstream tasks. Sun et al. \cite{sun2025} show that under extreme imbalance, objective choice can outweigh architectural differences in fraud detection performance. Similarly, Chen et al. \cite{chen2026} demonstrate that static model fit does not necessarily translate to superior dynamic hedging outcomes, underscoring the distinction between evaluation metrics and operational objectives.

Representation robustness has also been addressed in other domains. Tian and Wang \cite{tian2025} propose integrating dimensionality reduction and contrastive objectives directly into recommendation systems to denoise latent representations. In inverse design problems, Yuan et al. \cite{yuan2022} demonstrate that jointly trained forward-inverse neural architectures can mitigate instability caused by non-unique mappings. These works collectively suggest that structured refinement and objective-aware design can improve robustness without necessarily increasing architectural complexity.

Inspired by these perspectives, our work introduces temporal aggregation as a lightweight yet effective signal-level stabilization mechanism, bridging sentiment classification and time-series anomaly detection through objective alignment rather than architectural expansion.

\section{Problem Formulation}

We represent a dataset of timestamped user comments as:
\begin{equation}
D = \{(x_i, t_i)\}_{i=1}^N,
\end{equation}
where $x_i$ represents the text of the comment and $t_i$ is its timestamp. User comments are temporally ordered and may be generated at non-uniform intervals. We focus on creating aggregated sentiment scores within fixed windows that reveal temporal trends. 
We define time windows $T_k$ using either:
\begin{itemize}
\item Count-based segmentation: every $n$ comments form a window; or

\item Time-based segmentation: windows span fixed elapsed time intervals (e.g., daily).

\end{itemize}
Let $T_k$ = \{$x_i$|$t_i$ falls within window$k$\}. For each comment $x_i$, we obtain a sentiment prediction $y_i$ from a fine-tuned RoBERTa model. The predictions use three sentiment classes: positive, neutral, and negative. These class labels are mapped to numerical scores:
\begin{equation}
S(T_k) = \frac{1}{|T_k|} \sum_{i \in T_k} s_i,
\end{equation}
where $s_i \in \{-1, 0, +1\}$ denotes negative, neutral, and positive predictions from a RoBERTa model. The change between windows is computed as:
\begin{equation}
\Delta S(T_k) = S(T_k) - S(T_{k-1}),
\end{equation}
and significant downward changes indicate potential anomalies.

\section{Methodology}

\subsection{RoBERTa as a Semantic Feature Extractor}
We employ RoBERTa, a robustly optimized transformer-based language model, as the backbone of our sentiment classification pipeline. RoBERTa improves upon BERT through optimized training procedures and dynamic masking, yielding stronger contextual representations, especially on short, informal text typical of social media and user feedback.

We fine-tune the pretrained RoBERTa model on labeled sentiment datasets representative of our task domain. During fine-tuning, we use a classification head with three output logits corresponding to the sentiment classes. Cross-entropy loss and standard optimization techniques (e.g., AdamW) are used to train the model.

\subsection{Temporal Aggregation Framework}
Once per-comment sentiment scores are obtained, they are aggregated within fixed windows to smooth out noise. The choice of window size impacts both the stability and sensitivity of the resulting time series: larger windows reduce variance but may dilute rapid changes, whereas smaller windows capture transient shifts but may be too noisy.

We evaluate both count-based and time-based windows in our experiments and analyze their effects on the quality of temporal signals.

\subsection{Topic-Aware Sentiment Aggregation}

While global sentiment aggregation captures overall feedback trends, it does not distinguish between different operational causes of negative feedback. In practice, abnormal sentiment drops may arise from multiple sources such as flight delays, baggage issues, or customer service problems.

To preserve semantic information, we extend the aggregation framework with a topic-aware component. Each comment is associated with a topic label derived from the complaint category in the dataset (e.g., \textit{Late Flight}, \textit{Cancelled Flight}, \textit{Customer Service Issue}, \textit{Lost Luggage}). These topic labels serve as aspect indicators representing operational dimensions of user feedback.

Sentiment aggregation is then performed separately for each topic. Let $z$ denote a topic label and $T_k^z$ represent the subset of comments belonging to topic $z$ within time window $k$. The topic-specific aggregated sentiment is defined as

\begin{equation}
S_z(T_k) = \frac{1}{|T_k^z|} \sum_{i \in T_k^z} s_i
\end{equation}

This formulation produces a sentiment trajectory for each operational topic, enabling anomaly detection not only at the global level but also within individual feedback categories. Such topic-level monitoring provides more interpretable and actionable insights for operational teams.

\subsection{Anomaly Detection Mechanism}
Our anomaly detection approach is based on scoring changes between adjacent time windows. Assuming normal sentiment progression exhibits small random fluctuations around a baseline, significant negative shifts indicate departures from normal behavior. We set an anomaly threshold $\tau$ based on historical changes, such as 
\begin{equation}
\tau = \mu_{\Delta S} - \alpha \sigma_{\Delta S},
\end{equation}
where $\mu_{\Delta S}$ and $\sigma_{\Delta S}$ denote the mean and standard deviation
of historical sentiment changes, and $\alpha$ controls detection sensitivity.

The detection mechanism is interpretable and straightforward, suitable for integration with monitoring dashboards and alerting systems.

\section{Implementation}

\subsection{System Architecture}
The implementation comprises four connected modules:
\begin{itemize}
  \item Preprocessing Module: Cleans text and normalizes timestamps.
  \item RoBERTa Sentiment Predictor: Applies the fine-tuned RoBERTa model to generate per-comment sentiment labels.
  \item Temporal Aggregator: Segments data into windows and computes aggregated scores.
  \item Anomaly Detector and Visualizer: Computes changes, detects signals, and outputs visual summaries.
\end{itemize}

\subsection{Data Preprocessing}
Raw comment data were parsed and cleaned (timestamp normalization, duplicate removal, and basic text validation) before sentiment inference.

\subsection{RoBERTa Fine-Tuning and Inference}
We fine-tuned a roberta-base model with a three-class classification head using standard hyperparameters (batch size 32, learning rate $2\times10^{-5}$, AdamW optimizer, 3 epochs).

For inference, input text was tokenized using the RoBERTa tokenizer with padding and truncation to a maximum sequence length of 128 tokens. Predicted class labels were mapped to numerical scores as defined in Section 3.

\subsection{Temporal Windowing and Aggregation}
We experimented with two strategies:

Count-Based Windows

Each window comprised 100 successive comments, ensuring equal sample size per window.

Time-Based Windows

Windows represented fixed calendar periods (e.g., daily), aligning with natural reporting intervals.

For each window, aggregated sentiment scores $S(T_k)$ were computed via mean of individual scores.

\subsection{Anomaly Detection Thresholding}
We computed first-order differences $\Delta S(T_k)$ for the entire series. The anomaly detection threshold was set using a robust statistical approach:
\begin{equation}
\tau = \mu_{\Delta S} - 1.5 \sigma_{\Delta S},
\end{equation}
where $\mu_{\Delta S}$ and $\sigma_{\Delta S}$ denote the mean and standard deviation of historical window-to-window sentiment changes, respectively.

A window $T_k$ is flagged as anomalous if:
\begin{equation}
\Delta S(T_k) < \tau.
\end{equation}

\section{Experiments}

\subsection{Dataset Characteristics}
We applied the framework to a dataset of user comments collected from a public social media platform. The dataset contained tens of thousands of comments spanning several months, exhibiting informal language, emojis, and class imbalance with a predominance of negative feedback typical of real user discussion.

\subsection{Evaluation Metrics}
We assessed:
\begin{itemize}
  \item \textbf{Aggregate Score Variance:} Measures stability of sentiment scoring.
  \item \textbf{Anomaly Detection Sensitivity:} Counts and contextual relevance of flagged windows.
  \item \textbf{Computational Efficiency:} Average inference time per comment.
\end{itemize}

\section{Results}

\subsection{Overall Sentiment Distribution}

We first analyze the overall sentiment distribution predicted by the RoBERTa classifier. The distribution of sentiment scores across the dataset is as follows:

\begin{itemize}
\item Negative sentiment: 47.19\%
\item Neutral sentiment: 31.02\%
\item Positive sentiment: 21.79\%
\end{itemize}

The results indicate that the dataset is dominated by negative sentiment, which is consistent with the nature of airline-related customer feedback where users tend to express dissatisfaction or complaints. This distribution provides a realistic test environment for evaluating anomaly detection methods in negative-feedback-heavy scenarios.

\subsection{Temporal Sentiment Trajectory}

\begin{figure}[htbp]
\centerline{\includegraphics[width=\columnwidth]{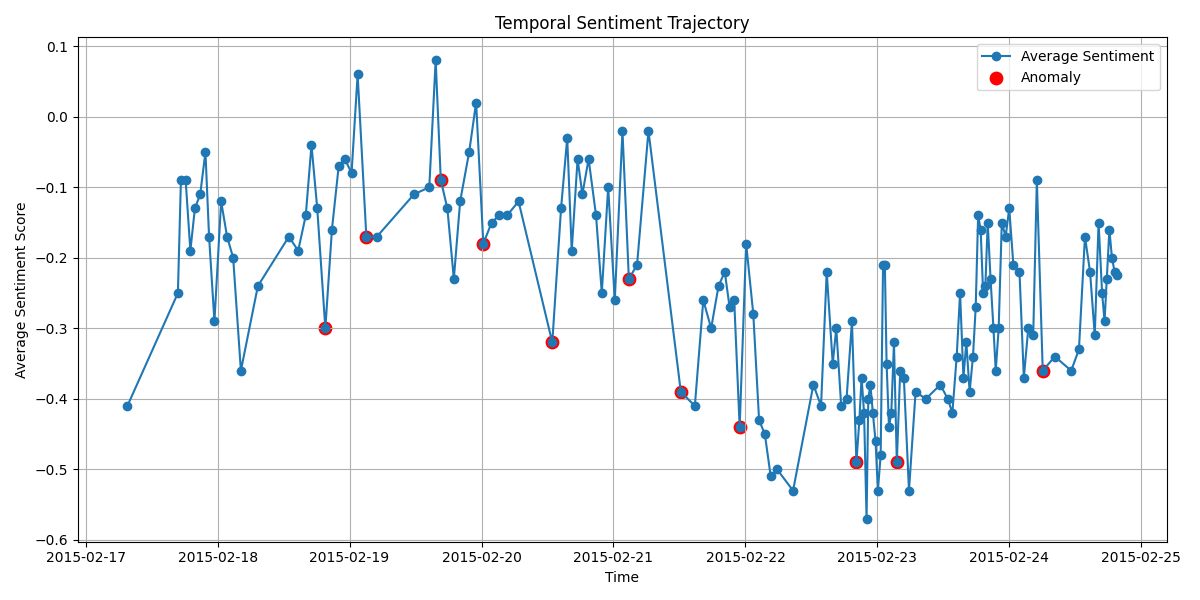}}
\caption{Temporal aggregated sentiment trajectory. Red markers indicate detected anomaly windows where significant sentiment drops occur.}
\label{fig:sentiment_traj}
\end{figure}

Fig.~\ref{fig:sentiment_traj} illustrates the temporal evolution of aggregated sentiment scores using count-based windows of 100 comments each. The aggregated sentiment scores range approximately from $-0.57$ to $0.08$, demonstrating noticeable fluctuations over time.

The first ten aggregated sentiment scores range from -0.41 to -0.05,
indicating moderate variability during the initial monitoring period.

These results indicate that sentiment dynamics are non-stationary and exhibit both gradual changes and sudden drops. Several abrupt downward movements can be observed, suggesting potential abnormal feedback episodes.

\subsection{Anomaly Detection Based on Sentiment Changes}

To detect abnormal sentiment behavior, we compute the first-order difference of aggregated sentiment scores:

\begin{equation}
\Delta S(T_k) = S(T_k) - S(T_{k-1}).
\end{equation}

An anomaly threshold is determined using:

\begin{equation}
\tau = \mu_{\Delta S} - 1.5\sigma_{\Delta S},
\end{equation}

which yields $\tau = -0.1693$ in our experiments.

\begin{figure}[htbp]
\centerline{\includegraphics[width=\columnwidth]{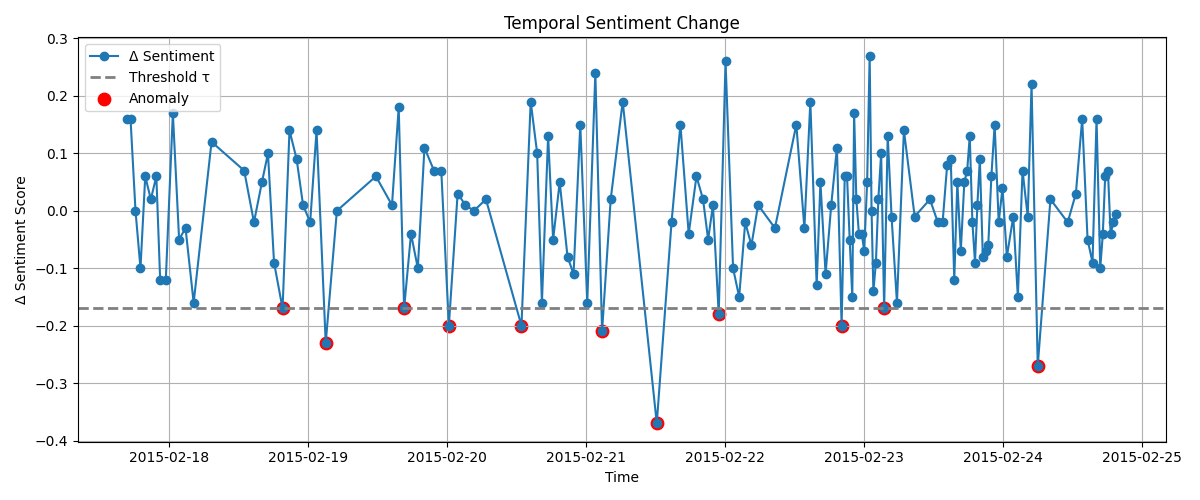}}
\caption{Temporal sentiment change ($\Delta$Score) with anomaly detection threshold $\tau = -0.1693$. Downward spikes crossing the threshold correspond to abnormal sentiment deterioration.}
\label{fig:delta_sentiment}
\end{figure}

Fig.~\ref{fig:delta_sentiment} visualizes the temporal sentiment changes and the anomaly detection threshold. Detected anomalies appear as sharp downward spikes that cross the threshold line, providing an intuitive interpretation of abnormal sentiment deterioration.

Using this criterion, eleven anomalous windows are detected at indices:

\begin{center}
$\{20, 26, 31, 37, 42, 54, 57, 65, 81, 98, 132\}$.
\end{center}

\subsection{Before--After Sentiment Comparison}

To further validate anomaly detection results, Table~\ref{tab:anomaly_comparison} compares sentiment scores before and after each detected anomaly window. The results show that anomalies consistently correspond to abrupt sentiment drops rather than persistently low sentiment values.

For example, window 57 exhibits the largest sentiment decline from $-0.02$ to $-0.39$, corresponding to a drop of $-0.37$. Similar sudden declines are observed across other anomaly windows, confirming the effectiveness of the proposed detection method in capturing unexpected sentiment deterioration.

\begin{table}[htbp]
\caption{Before--After Sentiment Comparison for Detected Anomalies}
\begin{center}
\begin{tabular}{|c|c|c|c|}
\hline
\textbf{Window} & \textbf{Previous Score} & \textbf{Current Score} & \textbf{$\Delta$ Score} \\
\hline
20  & -0.13 & -0.30 & -0.17 \\
26  & 0.06  & -0.17 & -0.23 \\
31  & 0.08  & -0.09 & -0.17 \\
37  & 0.02  & -0.18 & -0.20 \\
42  & -0.12 & -0.32 & -0.20 \\
54  & -0.02 & -0.23 & -0.21 \\
57  & -0.02 & -0.39 & -0.37 \\
65  & -0.26 & -0.44 & -0.18 \\
81  & -0.29 & -0.49 & -0.20 \\
98  & -0.32 & -0.49 & -0.17 \\
132 & -0.09 & -0.36 & -0.27 \\
\hline
\end{tabular}
\label{tab:anomaly_comparison}
\end{center}
\end{table}

\subsection{Semantic Analysis of Anomalous Windows}

To investigate whether detected anomalies correspond to coherent complaint patterns, we analyze the distribution of negative reasons within anomalous and normal windows.

\begin{figure}[htbp]
\centerline{\includegraphics[width=\columnwidth]{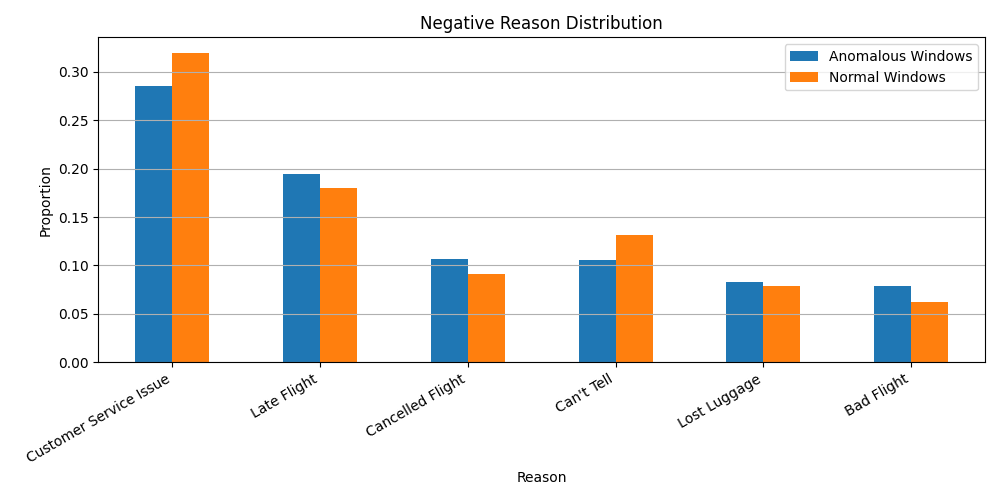}}
\caption{Comparison of negative complaint reason distributions between anomalous and normal windows. Anomalous windows exhibit stronger concentration of delay and service-related complaint categories.}
\label{fig:reason_distribution}
\end{figure}

Fig.~\ref{fig:reason_distribution} presents the comparative distribution of major complaint categories. Several observations can be made:

\begin{itemize}
\item Customer service issues remain the most dominant complaint type in both anomalous and normal windows.
\item Anomalous windows exhibit slightly higher proportions of late flight and cancelled flight complaints.
\item Flight attendant complaints and service-related issues also show increased concentration within anomalous windows.
\end{itemize}

These findings suggest that sentiment drops are not random fluctuations but are associated with structured complaint topics. This provides semantic evidence supporting the validity of anomaly detection results. While we do not claim causal relationships, the consistent shifts in complaint category distributions provide strong supporting evidence that detected anomalies reflect meaningful changes in user feedback behavior.

\subsection{Topic-Level Sentiment Dynamics}

While global sentiment monitoring reveals overall feedback trends, it does not indicate the operational causes behind sentiment fluctuations. To address this limitation, we extend the analysis by examining sentiment trajectories for individual complaint topics.

\begin{figure}[htbp]
\centerline{\includegraphics[width=\columnwidth]{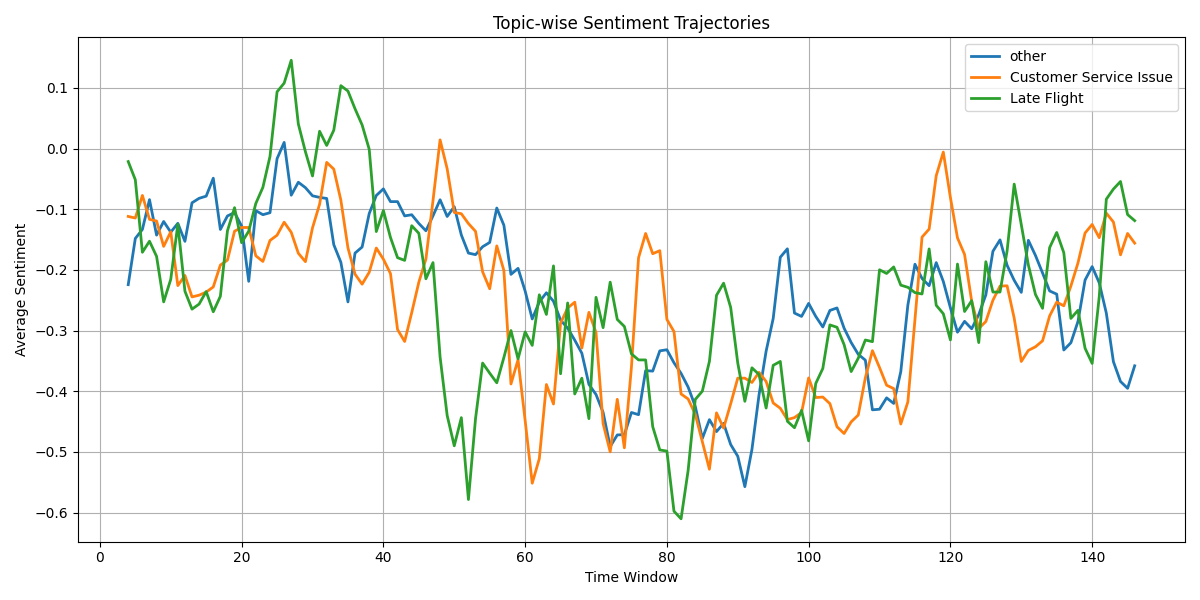}}
\caption{Topic-wise sentiment trajectories across time windows. Each curve represents the aggregated sentiment trend for a specific complaint category.}
\label{fig:topic_traj}
\end{figure}

Figure~\ref{fig:topic_traj} shows the sentiment trajectories for several major complaint categories, including \textit{Customer Service Issue}, \textit{Late Flight}, and \textit{Other}. The trajectories represent aggregated sentiment values computed within each time window for comments belonging to the corresponding topic.

Compared with the global sentiment trajectory, topic-level monitoring reveals more detailed behavioral patterns. For example, sentiment associated with \textit{Late Flight} exhibits noticeable fluctuations across several windows, indicating periods where flight delays generated clusters of negative feedback. Similarly, the \textit{Customer Service Issue} trajectory demonstrates localized sentiment deterioration during specific intervals.

These observations suggest that sentiment anomalies are not uniformly distributed across all topics but are often concentrated within specific operational categories.

To further improve interpretability of multi-topic sentiment dynamics, we visualize the sentiment evolution across topics using a heatmap representation.

\begin{figure}[htbp]
\centerline{\includegraphics[width=\columnwidth]{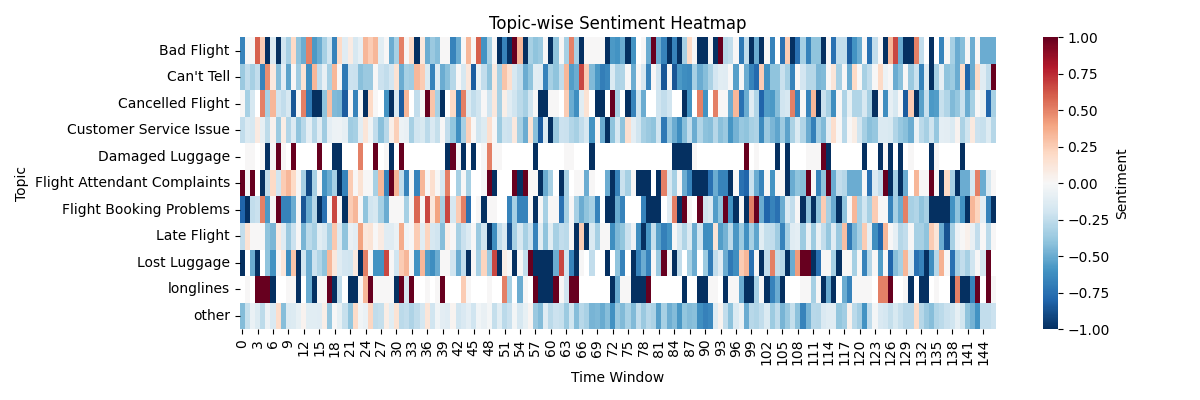}}
\caption{Topic-wise sentiment heatmap over time windows. Colors represent aggregated sentiment scores for each complaint topic.}
\label{fig:topic_heatmap}
\end{figure}

Figure~\ref{fig:topic_heatmap} presents a topic-time sentiment heatmap where rows correspond to complaint categories and columns represent time windows. Color intensity indicates the aggregated sentiment score, with red representing positive sentiment and blue representing negative sentiment.

This visualization allows rapid identification of abnormal sentiment clusters. For instance, several windows show concentrated negative sentiment within the \textit{Lost Luggage} and \textit{Late Flight} categories, suggesting that operational disruptions related to baggage handling and flight delays contributed to the observed sentiment deterioration.

Compared with the single aggregated sentiment signal, topic-aware visualization provides more diagnostic insights into the underlying causes of abnormal feedback patterns.

\subsection{Summary of Findings}

Overall, the experimental results demonstrate the following.

\begin{itemize}
\item Temporal aggregation produces stable and interpretable sentiment trajectories.
\item Sudden sentiment drops can be effectively detected using first-order difference analysis.
\item The detected anomalies correspond to coherent complaint topics rather than noise.
\item Topic-aware sentiment monitoring reveals that anomalies are often concentrated within specific complaint categories such as flight delays and baggage issues, providing more interpretable insights into the operational causes of abnormal feedback patterns.
\end{itemize}

These findings directly support our core claims: temporal aggregation stabilizes noisy sentiment signals, change-based detection identifies abnormal feedback episodes, and semantic analysis confirms that detected anomalies correspond to structured complaint behavior rather than random fluctuations.

\section{Discussion}

Aggregated sentiment scores derived from RoBERTa predictions provide stable and informative signals suitable for temporal monitoring. The choice of window size and strategy affects noise and sensitivity: larger windows smooth noise but may overlook short-lived anomalies; smaller windows increase sensitivity but introduce volatility. RoBERTa’s contextual understanding contributes to meaningful score trends by capturing semantic nuances that simpler models might miss, albeit at increased computational cost.

The anomaly threshold parameter $\tau$ is tunable and can be adjusted according to the desired balance between false positives and detection sensitivity. In practice, combining sentiment trends with other signals such as volume spikes or topic shifts may further enhance reliability.

\section{Threats to Validity}

Several factors may affect the validity of our findings. First, sentiment predictions rely on a pretrained RoBERTa model, which may exhibit biases or misclassifications on domain-specific language such as sarcasm or implicit complaints. While aggregation mitigates individual prediction noise, systematic model bias may still influence aggregated scores.

Second, the choice of window size directly impacts anomaly sensitivity. Although we analyze reasonable window configurations, different applications may require tuning based on data volume and monitoring objectives.

Finally, our evaluation focuses on a single real-world dataset. While the observed patterns are consistent with known characteristics of customer feedback streams, future work should validate the approach across additional domains and platforms.

\section{Conclusion}

We have presented a temporal sentiment aggregation framework using a pretrained RoBERTa model to detect abnormal patterns in user feedback. Aggregated sentiment scores over time windows reveal meaningful trends and support practical anomaly detection, addressing challenges inherent in noisy individual text classification. Through empirical evaluation and implementation details, we demonstrate the utility of this approach for real-world monitoring applications and provide insights for model choice and system design.

In addition to global sentiment monitoring, our experiments demonstrate that topic-aware sentiment aggregation can provide more diagnostic insights into abnormal feedback patterns. By tracking sentiment trajectories across operational complaint categories, the system is able to identify not only when anomalies occur but also which underlying issues contribute to them.


\end{document}